\documentclass[10pt,twocolumn,letterpaper]{article}

\usepackage[pagenumbers]{cvpr} % To force page numbers, e.g. for an arXiv version

\usepackage{graphicx}
\usepackage{amsmath}
\usepackage{amssymb}
\usepackage{booktabs}
\usepackage{makecell}
\usepackage{multirow}
\usepackage[pagebackref,breaklinks,colorlinks]{hyperref}

\usepackage[capitalize]{cleveref}
\crefname{section}{Sec.}{Secs.}
\Crefname{section}{Section}{Sections}
\Crefname{table}{Table}{Tables}
\crefname{table}{Tab.}{Tabs.}

\def\cvprPaperID{1826} % *** Enter the CVPR Paper ID here
\def\confName{CVPR}
\def\confYear{2023}

\begin{document}

\title{Towards Arbitrary Text-driven Image Manipulation via Space Alignment}

% \and
\author{%
  Yunpeng Bai$^{1}$, Zihan Zhong$^{1}$, Chao Dong$^{2,3}$, Weichen Zhang$^{1}$, Guowei Xu$^{1}$,  Chun Yuan$^{1,4}$ \\[0.5em]
  $^{1}$ Tsinghua University, $^{2}$Shenzhen Institutes of Advanced Technology, Chinese Academy of Sciences\\ $^{3}$ Shanghai AI Laboratory, China, $^{4}$Peng Cheng Laboratory, Shenzhen, China \\[0.3em]
%   {\small \texttt{\{chenh, bohe, hywang66,  yxren, abhinav\}@umd.edu, sernamlim@fb.com} }
}

% \name{Yunpeng Bai$^{1}$ \qquad Chao Dong$^{2,3}$ \qquad Cairong Wang$^{1}$ \qquad Chun Yuan$^{1,4}$}
% \address{$^1$ Shenzhen International Graduate School, Tsinghua University, Shenzhen, China  \\
% $^2$Shenzhen Institutes of Advanced Technology, Chinese Academy of Sciences \\
% $^3$Shanghai AI Laboratory, China \qquad $^4$ Peng Cheng Laboratory, Shenzhen, China \\
%  byp20,wcr20@mails.tsinghua.edu.cn; chao.dong@siat.ac.cn; yuanc@sz.tsinghua.edu.cn}

\twocolumn[{%
\renewcommand\twocolumn[1][]{#1}%
\maketitle

% [width=0.97\textwidth,height=0.97\textwidth]
\begin{center}
\centering
\includegraphics[width=0.97\textwidth]{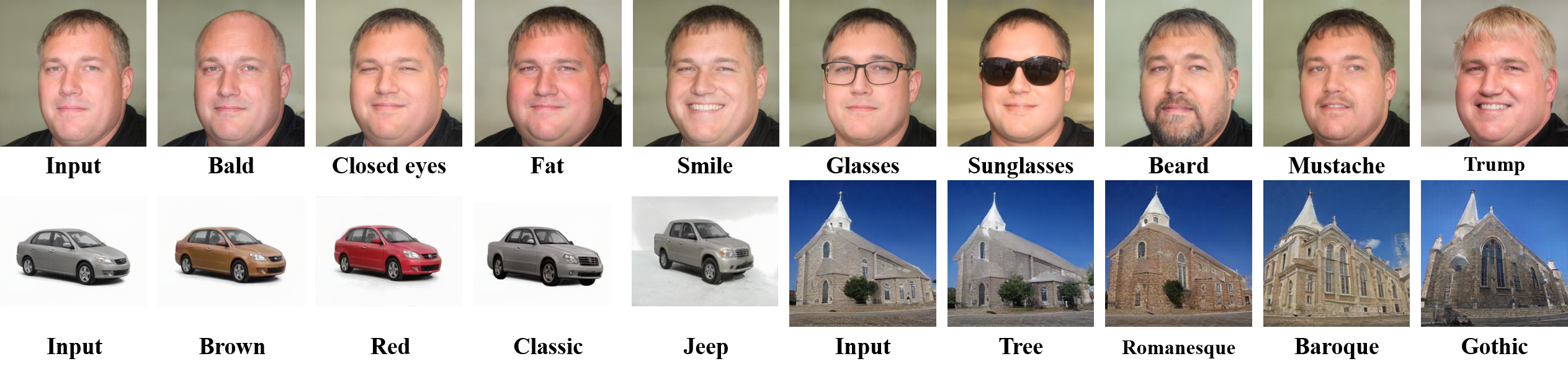}
\captionof{figure}{Overview of image manipulation results of the proposed method. Our method allows editing images with arbitrary text input and distinguishes between detailed attributes such as ``glasses'' and ``sunglasses'', ``beard'' and ``mustache''.}
\label{fig:overall}
\end{center}}]

% due to the relatively limited semantic discovery in the StyleGAN space, how to perform the desired attribute control on the latent code remains to be further explored.

%%%%%%%%% ABSTRACT
\begin{abstract}
The recent GAN inversion methods have been able to successfully invert the real image input to the corresponding editable latent code in StyleGAN.
By combining with the language-vision model (CLIP), some text-driven image manipulation methods are proposed. However, these methods require extra costs to perform optimization for a certain image or a new attribute editing mode. To achieve a more efficient editing method, we propose a new Text-driven image Manipulation framework via Space Alignment (TMSA). The Space Alignment module aims to align the same semantic regions in CLIP and StyleGAN spaces. Then, the text input can be directly accessed into the StyleGAN space and be used to find the semantic shift according to the text description. The framework can support arbitrary image editing mode without additional cost. Our work provides the user with an interface to control the attributes of a given image according to text input and get the result in real time. Extensive experiments demonstrate our superior performance over prior works. 
%最近的gan inversion的方法能够很好地建模，但是还有待于探索
%现有的方法都过于繁琐

\end{abstract}

%!!!!!!!!!考虑在哪里加上这样一句话 就是clip的空间是用余弦相似度训练的，styleGAN是早就被观察到具有这样的性质，为两个空间的对齐提供了一种可能性。 所以两个空间是可以对齐的  
%还有就是 clip的空间是可以通过文本直接接入到对应的区域上的

%%%%%%%%% BODY TEXT
\section{Introduction}
\label{sec:intro}
Generative Adversarial Networks (GANs)\cite{goodfellow2020generative} are the current mainstream generative models, and some of them have shown great power for high-quality image synthesis, such as StyleGANs\cite{karras2019style}.
% represented by the recent StyleGANs have extensive power for high-quality image synthesis. 
% In addition, the latent space of StyleGAN was found to exhibit disentanglement properties and encode rich semantics. This property has facilitated the emergence of these researches on image manipulation based on the StyleGAN inversion. 
In addition, it is observed that the latent space of GAN exhibits disentanglement properties and encodes rich semantics. This foundation has promoted the research on image manipulation based on GAN inversion. These image manipulation methods enable semantic editing of existing photographs.
%图像操作的应用场景

The common practice is to first invert the input image into a latent code, and then edit the latent code to manipulate the corresponding attributes, such as color and shape. For the first step, recent high-quality \cite{tov2021designing,alaluf2021restyle,wang2022high,roich2022pivotal}
inversion schemes have already been able to successfully invert the real image input to the corresponding latent code, which can faithfully reconstruct the original image and is semantically editable.
% how to make the desired attribute control on the latent is still to be explored more, because the semantic discovery of StyleGAN space is relatively limited.
However, the next step, how to realize a perfect attribute control is still challenging. Because the latent codes are abstract representations, it is difficult to detect semantics in the latent space.
% the semantics of the latent code are strongly coupled together.
% Earlier methods either find these latent paths for editing in a supervised manner using annotated data, or find some meaningful directions first in an unsupervised manner and then manually annotated them to determine the semantic meaning. 

Earlier methods can be divided into two groups. The first group finds the latent paths for editing in a supervised manner using annotated data\cite{shen2020interpreting,abdal2021styleflow}. The second one finds some meaningful directions first in an unsupervised manner and then manually annotates them to determine the semantic meaning\cite{harkonen2020ganspace,shen2021closed,voynov2020unsupervised}. 
If users use these methods to find the desired direction for editing, further manual efforts or extensive annotated data is required. Based on the aforementioned works, other researchers have further explored manipulation directions using pre-trained 3DMM \cite{tewari2020stylerig} and normalized flow \cite{abdal2021styleflow}.
% other researchers have further explored the manipulation directions. Tewari et al [37] utilize pre-trained 3DMM to edit expressions, poses and face illumination. Abdal et al [3] use a normalized flow to do a better de-entanglement on latent code. 
However, these methods still only support some predefined semantic directions for image manipulation, and cannot achieve the specific attribute editing effect that users desire.

% In order to make image manipulation no longer limited to preset semantic directions and additional manual effort to discover new controls, 
The above image manipulation works are limited in preset semantic directions and require additional manual efforts to new controls. To overcome this limitation, 
% recent works combine Contrastive Language-Image Pre-training (CLIP) models to develop text-driven image manipulation methods.
% StyleCLIP exploits the ability of natural language to represent rich visual concepts and CLIP's language-vision embedding space to provide multi-modal supervision, resulting in amazing image editing results. 
StyleCLIP\cite{patashnik2021styleclip} exploits the ability of CLIP\cite{radford2021learning} and develops text-driven image manipulation methods. However, in order to edit a certain image or to mine a new manipulation direction, StyleCLIP needs extra costs.
% However, for each new attribute described by the text, it is required to retrain a model or optimize an image for a few minutes and adjust the control parameters. 
% The global direction method in StyleCLIP also requires tedious construction process and has limited effect. 
CLIP2StyleGAN\cite{abdal2022clip2stylegan} uses CLIP's text encoder as multi-attribute classifier to label semantic controls in an unsupervised way.
However, CLIP2StyleGAN can only support a few 
attribute manipulation modes.
% The whole process is also very complicated, such as labeling each direction, classifying codes in StyleGAN space with SVM, and still obtaining a limited variety of attribute editing directions. %但是整个流程仍然需要复杂的流程却仍很难获得某个想要的属性编辑方向 %但是他们对于StyleGAN空间的探索仍然是有限的,这导致了每想要编辑一个新属性,就需要重新训练一个模型或者对一张图片优化几分钟的时间以及调节控制参数

To address these issues, we propose a new text-driven image manipulation framework called \textbf{TMSA}.
Our method includes a Space Alignment module, the goal of which is to align the same semantic regions in CLIP and StyleGAN spaces, and then allow arbitrary text input to be directly accessed into the StyleGAN space. 
% TediGAN\cite{xia2021tedigan} is a similar implementation, which requires text-image paired data for training, making them limited to face images and requiring empirical style-mixing operations. 
% However, we utilize CLIP's multi-modal embedding space and only use the image data to learn a mapping that can also maps the text features to StyleGAN's space. 
By combining with existing inversion methods, 
our framework provides the user with an interface to control the attributes of the edited result with arbitrary text input and obtain the result in real time without much adjustments. Our method can support arbitrary image editing mode without additional cost. As shown in Figure \ref{fig:overall}, our methods can support various detailed and abstract attributes editing.

\begin{figure*}[t]
    \centering
    \includegraphics[width=\linewidth]{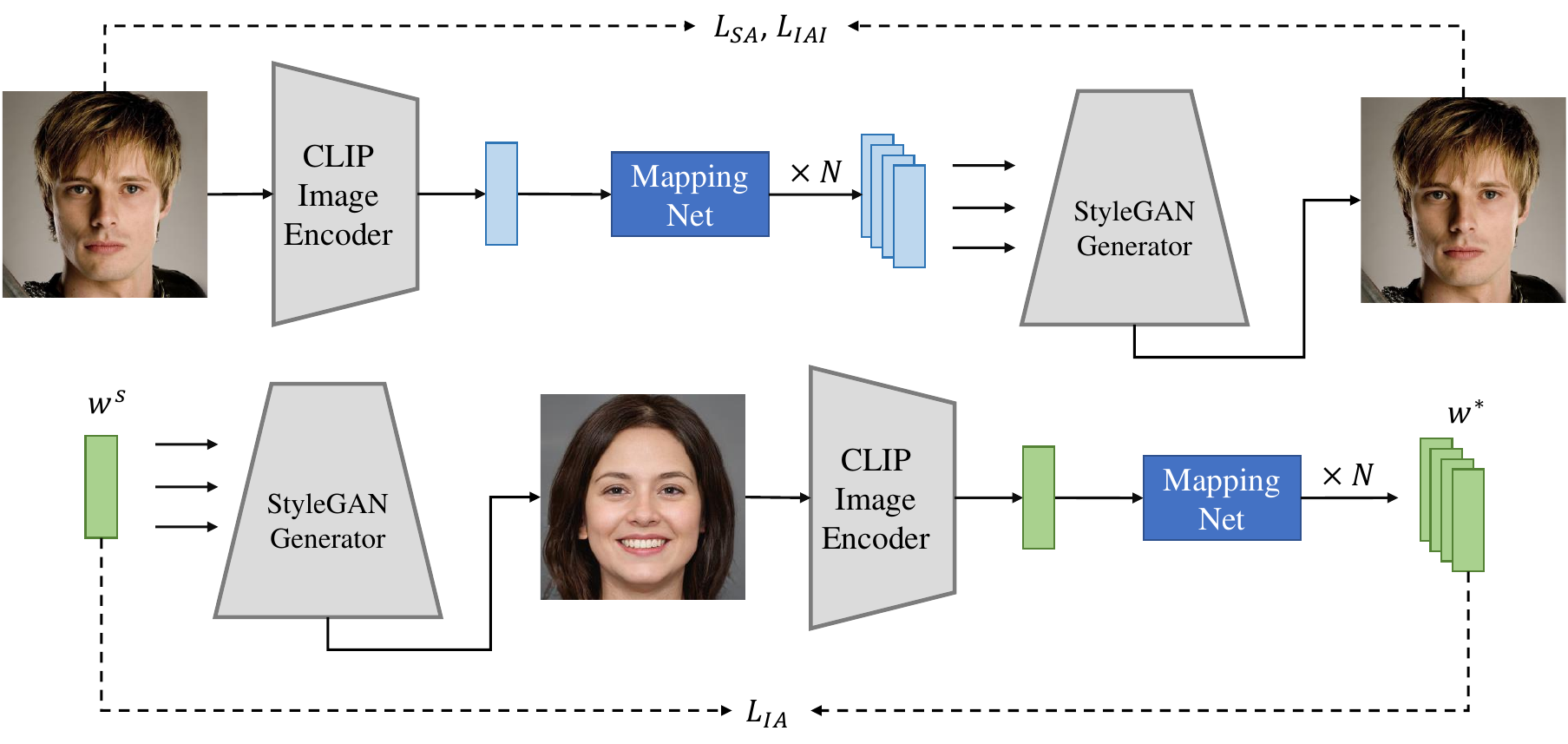}
    \caption{ The proposed Space Alignment module. During training we map the image embeddings from the CLIP's image encoder to latent code in the StyleGAN space, and then generate images with the same attributes as the original images through the generator. We also sample some generated latent $w^s$ to enhance the accuracy of space alignment}
    \label{fig:arch1}   
    % \vspace{-0.1in}
\end{figure*}

\section{Related Works}
\label{sec:related}

%写的时候可以参考其他文章的related work部分
\subsection{Image Manipulation in Latent Space}
%这里写基于ganinversion图像编辑的方式是怎么做的 介绍一些比较基本的方法，比如interfaceGAN，e4e什么的
The latent space of StyleGAN\cite{karras2019style} has shown great potential in representing the semantics of images, motivating many works to achieve image manipulation in the latent space. In order to utilize the power of pre-trained GANs for editing real images, it is vital to invert a given image back into the latent space close to the pre-trained GANs'. These inverted codes are expected to be semantically editable to achieve the goal of image manipulation. This process is often referred to as GAN inversion. Existing GAN inversion schemes can be roughly categorized into three classes: optimization-based\cite{abdal2019image2stylegan,abdal2020image2stylegan++}, encoder-based\cite{tov2021designing,alaluf2021restyle,wang2022high}, and hybrid approaches\cite{zhu2020domain,roich2022pivotal}. e4e\cite{tov2021designing} is a represented encoder-based method that inverts the input images into a $W^+$ space close to the original $W$ space of StyleGAN, achieving a better trade-off between distortion and editability. Based on these approaches, some works have explored the space of StyleGAN and found some meaningful semantic editing paths. However, these methods all require manual effort or extensive annotated data.
% InterFaceGAN uses the latent semantics interpreted by the conditional technique, to control the facial attributes.
% As e4e is a strong inversion method that allows subsequent image editing, our method uses it to get meaningful latent codes of the images.
\subsection{Text-driven Image Manipulation}
%这里写基于文本的图像编辑 介绍下styleclip，hairclip clip2StyleGAN和clipstyler  注意要写下styleclip和clip2StyleGAN的缺点
Text-driven image manipulation aims at changing certain parts of the images according to the text descriptions. By utilizing the powerful text-image representations capacities of CLIP\cite{radford2021learning}, a few works based on it have been developed for text-driven image manipulation. StyleCLIP\cite{patashnik2021styleclip} is the first work to develop a text-based interface for StyleGAN imagery manipulation. It proposes three schemes to edit images using CLIP: latent optimization, latent mapper, and global directions. However, these three methods have certain drawbacks. The optimization scheme needs several minutes to edit an image. The latent mapper needs to train a specific mapper for every text prompt, which is unsuitable for real-time applications. The global direction also requires a tedious process and has limited effect.

Based on StyleCLIP, HairCLIP\cite{wei2022hairclip} develops a hair editing framework that supports different texts or reference images as conditions and achieves disentangled hair editing. Although it could partially address the drawbacks of StyleCLIP, it still needs to train different sub hair mappers and the semantic modulation module. CLIPStyler\cite{kwon2022clipstyler}, focusing on style transfer, proposes a lightweight network to match the styles of images and text conditions. However, it also requires retraining the model for each given text condition.

% However, it also suffers from the disability of real-time style transfer for the need of training the model with each given text condition.

CLIP2StyleGAN\cite{abdal2022clip2stylegan} is another pipeline that tries to link the pre-trained latent spaces of StyleGAN and CLIP without additional supervision. However, this unsupervised manner cannot find meaningful semantic directions as desired. Moreover, the procedure is tricky and slow, such as using the PCA algorithm to find directions of CLIP space and using SVM to classify codes in StyleGAN space. AnyFace\cite{sun2022anyface} is a similar implementation, which requires text-image paired data for training, making them limited to face images. 
Unlike the approaches above, our method develops a more efficient editing framework based on Space Alignment between CLIP and StyleGAN without a complex and time-consuming training procedure.
\subsection{Text-to-Image Generation}
%介绍下dall-e,最近的一些diffusion的工作，TediGAN, 再随便加两个比较新的Text-to-Image的论文
How to generate meaningful images from text input is also a popular topic, which needs powerful generative models with cross-modal understanding abilities. TediGAN\cite{xia2021tedigan} proposes a unified framework for text-to-image generation and manipulation. It designs a GAN inversion technique that can map multi-modal inputs into a common latent space of a pretrained StyleGAN. The text-image matching is achieved by mapping them into a common space.
DALL-E\cite{ramesh2021zero} proposes a simple two-stage approach for this task, which trains a discrete variational autoencoder to compress images first, and then uses an autoregressive transformer to decode images by learning the prior distribution over the text and image tokens. CogView2\cite{ding2022cogview2} proposes an approach based on hierarchical transformers and and local parallel autoregressive generation, which remedies the slow generation and high complexity of autoregressive models.

Recently, diffusion models have attracted extensive attention as they can generate high-fidelity images with great diversity. DALL-E2\cite{ramesh2022hierarchical} leverages the abilities of CLIP and diffusion models, using a diffusion prior to generate CLIP image embeddings based on a given text description and convert them to images using a diffusion decoder. Imagen\cite{saharia2022photorealistic} introduces a new diffusion architecture and diffusion sampling technique to generate images. It also utilizes the power of a large transformer to provide effective guidance for text understanding.
% Recently, diffusion models which could gain much diversity in generating images, and improves fidelity at the certain cost of diversity, have attracted extensive attention. 

% DALL-E2 leverages the abilities of CLIP and diffusion models, using a diffusion prior that generates a CLIP image embedding given a text caption, and a diffusion decoder to generate an image conditioned on the image embedding. 

\begin{figure*}[t]
    \centering
    \includegraphics[width=\linewidth]{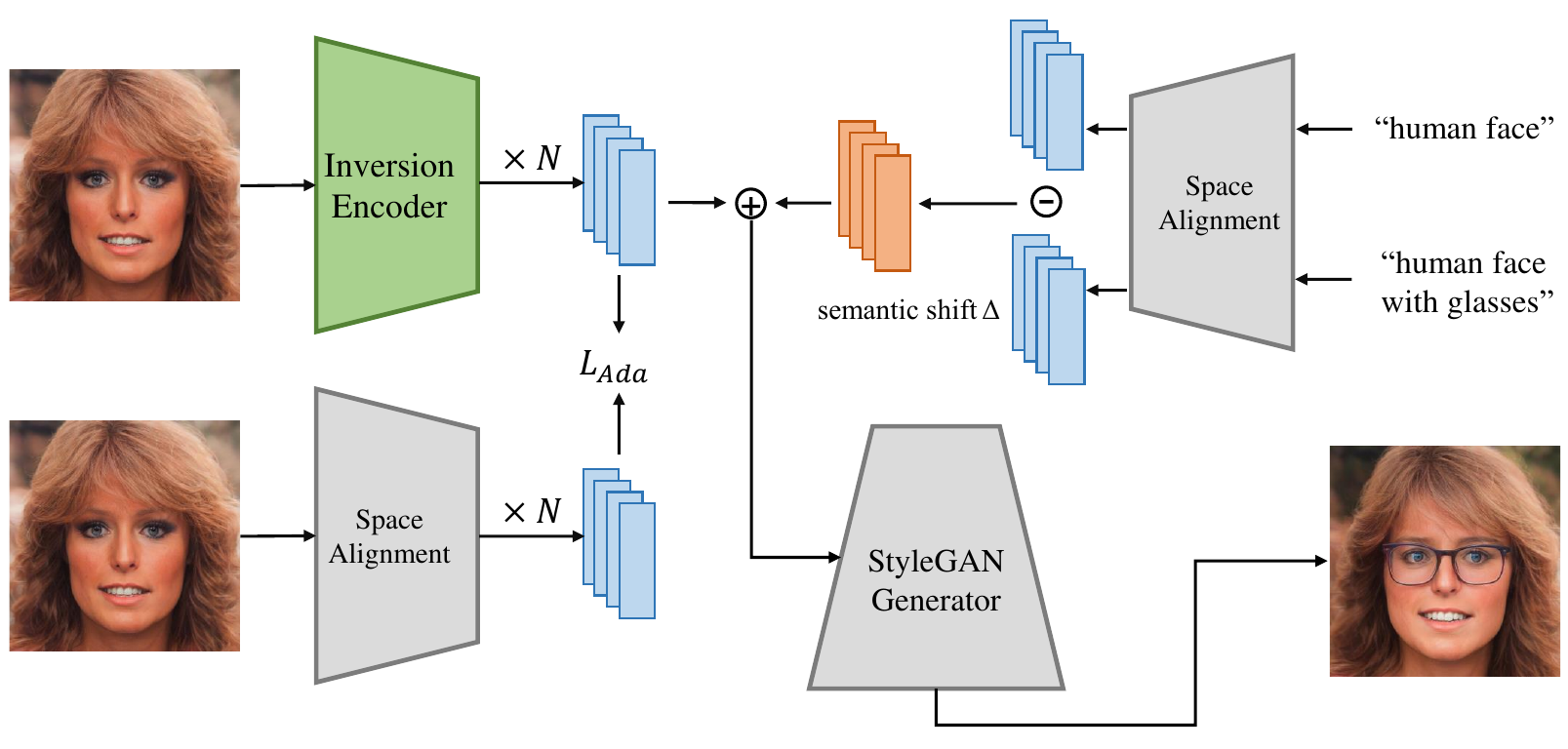}
    \caption{The process of combination with other inversion methods and arbitrary semantic shift extraction. Our method allows using two texts with different semantics to find the corresponding semantic shift in the space of StyleGAN. This shift is applied to edit all images without changing other attributes and identities.}
    \label{fig:arch2}   
    % \vspace{-0.1in}
\end{figure*}

\section{Proposed Method}
\label{sec:method}

Our image manipulation framework consists of several parts, an existing inversion encoder is used to obtain the latent code from the real image for editing, a space alignment module is used to find the semantic editing shifts corresponding to the text condition, and a pre-trained generator is used to generate images. Our space alignment module is trained using both real and generated images and can be adapted for specific inversion encoders to obtain better editing results.

%我们的编辑框架包含几个部分，一个已有的反演的编码器被用来从真实图像获得可以用来编辑的latent code，空间对齐模块用来找到文本所对应的语义编辑偏移，一个预训练好的生成器被用来生成图片。我们的空间对齐模块使用真实图像和生成图像训练，并且可以针对特定的反演编码器进行调整来获得更好的编辑的效果

\subsection{Space Alignment}
Since CLIP has been trained on an Internet-scale datasets, the encoder is able to extract various complex semantics from images and has been utilized in various zero-shot scenarios. The semantic information in its space is also rich and effective for text-driven image manipulation. We attempt to learn a mapping network to align this space with the space of pre-trained StyleGAN, so that the text embeddings from CLIP can match the latent representations from StyleGAN. 
% the text embeddings from CLIP can match latent representations with corresponding attributes in the space of StyleGAN. 
If the features of the text are directly mapped to the space of StyleGAN during training as TediGAN does, we need to use the texts paired with the image for supervision. Such text-image paired data is usually difficult to obtain. It is also hard to guarantee that text labels are matched with images. However, CLIP learns a joint language-vision embedding space, so we can use image instead of text to learn this mapping network. After training is finished, this network can also map the text features from CLIP space to the StyleGAN latent space.
% The learned spaces, although
% trained using only loose image-keyword pairing information, have been shown to be rich and effective for several
% zero-shot tasks, i.e., not requiring any further training or
% fine-tuning for new tasks. For example, in the context of
% image manipulation, CLIP and StyleGAN have been utilized for text guided editing [33] or zero-shot domain transfer [1
%由于clip在大规模的数据集上训练过，所以能够其encoder能够很好地从图像中抽取出各种复杂的语义，已有的研究将已经证明了clip的空间已经足够充分，以至于包含了StyleGAN空间中的语义信息，从而对于StyleGAN的image进行编辑。我们试图用一个映射网络将这个空间与训练好的StyleGAN的空间进行对齐，从而可以用文本的输入在StyleGAN的空间找到具有相应属性的latent表示。
%如果像TediGAN的做法将文本对应的特征映射到StyleGAN的空间中，就需要使用文本所对应的图像进行监督。而对于这种图像文本pair的数据，一方面数据标注困难，另一方面正如那句话所说，想用文本地标注通常很难描述出图像中的全部属性。但是，clip是一个图像文本joint的embedding空间，所以我们就可以只用图像数据来学习这个映射网络，从而同样可以将文本的特征通过该网络映射到StyleGAN的空间中。

We call such a process as Space Alignment, which is similar to the encoder-based GAN inversion but is completely different. The purpose of GAN inversion is to convert the given image back to the pre-trained GAN's latent space. The inversion code needs to be able to reconstruct the image faithfully. Space Alignment aims to align the regions with the same semantics in the two pre-trained networks' latent spaces, A and B. One of the spaces should contain most of the semantic information in the other space. After the space alignment is completed, we can use the region of an attribute in space A to find the region of the corresponding attribute in space B. Here, A and B represent CLIP space and StyleGAN space, respectively. 

The image embeddings encoded by the CLIP's encoder encode abstract attributes. It is unfeasible and unnecessary to do pixel-level image reconstruction with the CLIP embeddings. The training objective is to map such image embeddings to the latent code of the StyleGAN space. Then we can then generate images with the same attributes as the original image through the generator. The formulation of objective is:

\begin{equation}
\mathcal{L}_{SA}=1-\frac{E_I(I) \cdot E_I(I^*)}{\left\|E_I(I)\right\|_2 \cdot \left\|E_I(I^*)\right\|_2},
\end{equation}

where $E_I$ is the image encoder of CLIP. $I^* = G(w^*)$ is the image generated from the latent output of the mapping network. We also mapped the CLIP embeddings to a $W^+$ space of the StyleGAN as in previous works, 
% because the $W^+$ space was found to be semantically editable while maintaining image reconstruction fidelity, and was also easy to adapt to existing inversion methods. 
because the $W^+$ space is found to be semantically editable and easy to adapt to existing inversion methods. 
Therefore, $w^*$ is with the shape of $L \times C$, where $L$ is the number of layers and $C$ is the dimension of latent code. 
% meaning to have $L$ layers and each with a $C$-dimensional latent code.

%并且使用clip的的encoder抽取的图像特征中编码的是抽象的属性，用该特征完全地做像素级是不可行的，所以我们的目标是 图像输入进clip的encoder得到的特征映射到StyleGAN空间中的latent生成出的图片与原图具有相同的属性
%公式
%在这里面也是用clip的encoder来做属性的度量，即重建图和原图在clip空间中具有同样的余弦相似度

%也是到W+的空间，像之前的工作指出W+的空间更容易重建出属性，并且也容易与已有的inversion的方法相适应  W+空间类似于之前工作的定义 在这里面每一个w是代表StyleGAN中的对应的每一层的

\subsection{In-domain Adjustment}
%but also align the inverted code with the semantic knowledge encoded in the latent space为了能够保证  to map the image space to the latent space
% such that all codes produced by the encoder are in-domain. 
% fail to land the inverted code in the
% semantic domain of the original latent space.

%参考in domian gan是怎么写的 为了能够更好地适用于gan的空间 为了在latent上对图像进行对齐，也就是在latent上加监督
%在上面的过程中都是使用真实图像来进行训练的，并且只在图像级别上进行监督，这样并不能很好地保证将映射过来的latent放置在原来空间中对应的语义位置上 为了将align the inverted code with the semantic knowledge encoded in the latent space 对应到编码相应语义信息的区域上，我们又利用gan生成的图片对于该映射网络做了一个in-domian的调整。

In the above process, the training is conducted using real images in order to ensure that the mapping network can be applied to real images, and is supervised only at the image level. This does not guarantee that the mapped latent code is placed at the corresponding semantic location in the original space. In order to map the latent code to correct region, we perform an in-domain adjustment to the mapping network using the image generated by 
StyleGAN. We randomly sample a set of latent codes $w^s$ and then feed them into the generator $G(\cdot)$ to produce the corresponding synthetic image $I_{s}$. The embeddings of $I_{s}$ encoded by CLIP are then used as input to the mapping network. The corresponding latent codes $w_s$ are used as the supervision:

\begin{equation}
\mathcal{L}_{IA}=\sum_{i=1}^L \left\|w_i^*-w^s\right\|_2^2,
\end{equation}

where $w_i^*$ is the $i$-th layer of the mapped latent code, a $C$-dimensional code with the same shape as $w^s$. In addition, image level supervision is also added to the $I_{s}^*$ generated from $w^*$:

\begin{equation}
\mathcal{L}_{IAI}=1-\frac{E_I(I_{s}) \cdot E_I(I_{s}^*)}{\left\|E_I(I_{s})\right\|_2 \cdot \left\|E_I(I_{s}^*)\right\|_2},
\end{equation}

% land the inverted code in the
% semantic domain of the original latent space.

% a collection of latent codes z
% sam are randomly sampled and fed into
% G(·) to get the corresponding synthesis x
% syn
% % 然后这个映射网络使用clip编码的x的feature来作为输入，然后用对应的latent作为监督：

% Then, the encoder E(·) takes x syn and z
% sam as inputs and supervisions respectively and is trained with

% recover the input image at both the pixel level and the semantic level

\subsection{Arbitrary Semantic Shift Extraction}
Since CLIP is trained to obtain a space shared by text and image features, text features will land near image features with the same attributes in this shared space. After this mapping network is trained, we can also map text to the latent code with the corresponding attributes, just like an image. Thus, texts describing different attributes can be used to get semantic shifts directly in the StyleGAN space. In this way, we can avoid the complicated process of semantic direction construction and edit the attributes according to arbitrary text descriptions.

Taking face editing as an example, we can use some general text descriptions like ``human face'' and the text with some attributes like ``human face with glasses'' to access the StyleGAN space. These general descriptions will be mapped to an average face with certain attributes, and the identity of this average face is the same under different attributes. Therefore, we can use different such descriptions to get a global editing shift $\Delta w$ in the StyleGAN space, which can be used to edit all images without affecting other attributes and the identity.

% We also utilize the prompt engineering, which feeds several text descriptions with the same meaning to the text encoder and averages these embeddings as the final semantic embedding, to reduce the interference of text embedding noise. 
We also utilize the prompt engineering to reduce the interference of text embedding noise. Specifically, we feed several text descriptions with the same meaning to the text encoder and averages these embeddings as the final semantic embedding. Our prompt engineering is also implemented using the ImageNet prompt bank to decorate the text descriptions and obtain sentences with the same meaning.

% in order to obtain a stable direction in StyleGAN space.

\begin{figure}[t]
    \centering
    \includegraphics[width=\linewidth]{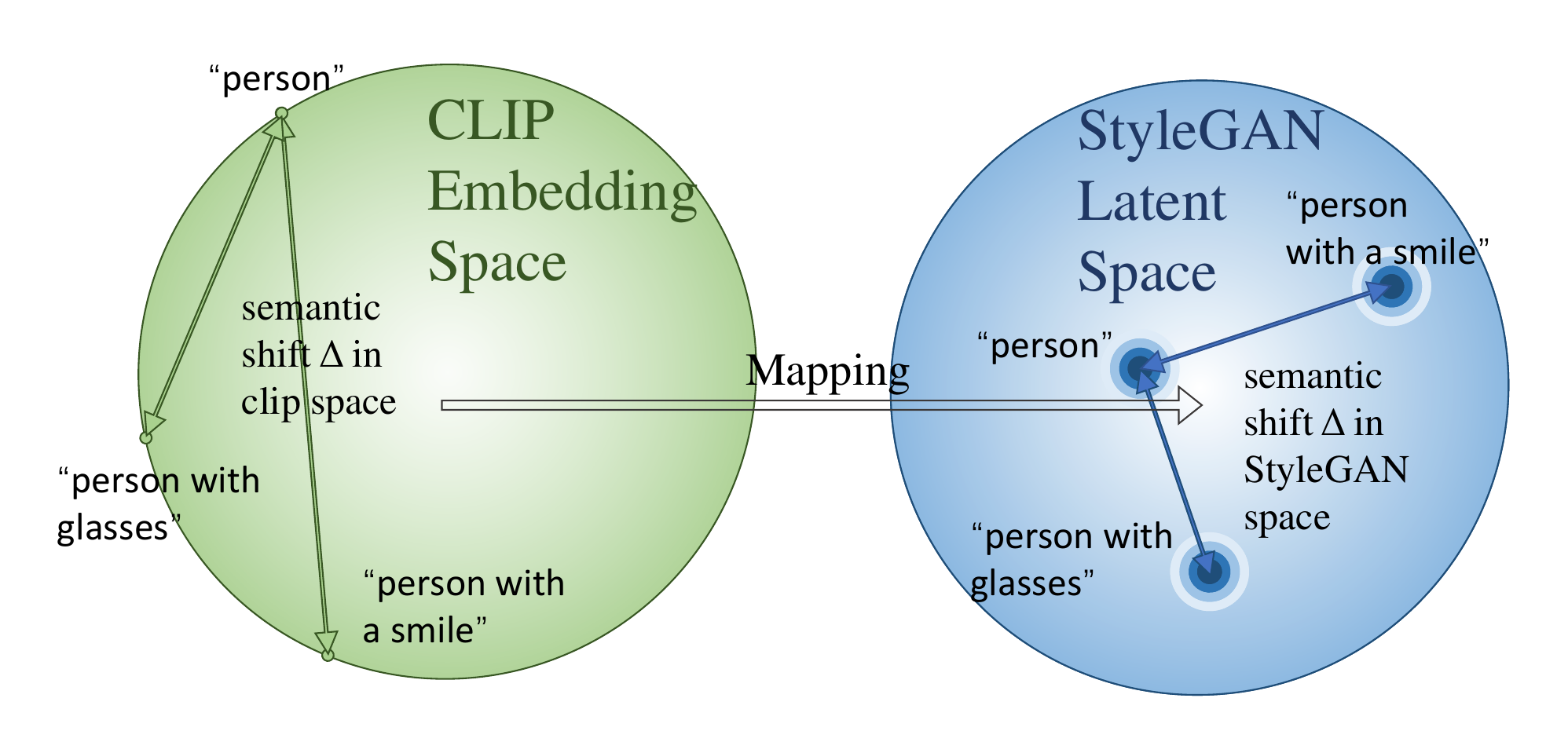}
    \caption{ Illustration of Space Alignment. Both the CLIP and StyleGAN spaces are well-trained and semantically decoupled. The CLIP space contains most of the semantics in the StyleGAN space, making it possible to align the same semantic regions in both spaces and to map semantic shift $\Delta$ in the CLIP space to the StyleGAN space. }
    \label{fig:shift}   
    % \vspace{-0.1in}
\end{figure}

\subsection{Adaption to Other Inversion Method}
Since this mapping network cannot reconstruct the image faithfully, in order to use it for image editing, it is necessary to combine it with the existing inversion methods and manipulate on the latent code from these methods. Although the $W^+$ space learned by our mapping network is somewhat different from that of the other inversion methods, the space semantics are basically aligned. It is possible to apply the semantic shift directly to the inverted codes obtained by the other methods. However, in order to achieve a better result, the network can be fine-tuned for a specific inverted space, as in the case of e4e\cite{tov2021designing}, we use their inverted code for fine-tuning:

%由于这个映射网络不能对输入图片进行重建，所以为了能够将其应用于图像编辑则还需要与已有的inversion方法相结合，在其他方法invert出来的latent的上进行编辑。虽然我们这里的mappingnet学习W+ganinversion方法的空间有一定的差异，但是空间语义基本是对齐的，所以可以直接在将这个语义的偏移应用在其他的gan inversion的方法上。但是为了达到一个更好的效果可以针对于某个特定的方法对网络进行微调，以e4e为例：

\begin{equation}
\mathcal{L}_{Ada}=\sum_{i=1}^L \left\|w_i^*-w_i^e\right\|_2^2,
\end{equation}

where $w^e$ is the inverted code from e4e encoder. Then, $G(w^e +\alpha \Delta w)$ can produce an edited image according to the attributes described by $\Delta w$, where $\alpha$ controls the manipulation strength. It is worth noting that our semantic shift here is different from the control direction defined in the previous work. Our shift is the distance of the average of different attribute codes, $\alpha$ set to 1 can be applied to most images to achieve corresponding attribute editing.
%值得注意的是我们这里的与之前工作只定义了一个方向不同,而是一个平均的属性shift,alpha设置为1可以应用到各种图像上

% 我们这里两个clip的空间和另一个StyleGAN的W+空间的对齐，但是和已有的方法的W+空间（比如e4e的空间）还是有一定的差别 为了更好利用已有的方法我们需要对针对其他的空间进行微调，对于同一张图片，同样的目的也是让相同语义区域的地方进行对齐

\section{Experiments}
\label{sec:experiments}
% In this section, we first introduce the experimental settings. Then we present qualitative and quantitative comparisons between the proposed method and several baseline models. Finally, we conduct several ablation studies to analyze the effect of our method.

\subsection{Implementation Details}
We conduct the experiments on several commonly used datasets, including the FFHQ dataset\cite{karras2019style}, the LSUN cars dataset\cite{yu2015lsun}, and the LSUN Church dataset\cite{yu2015lsun}. For each dataset, we use the corresponding pre-trained StyleGAN2\cite{karras2020analyzing} as the generator. The dimensions of the latent code are $18\times512$, $16\times512$ and $14\times512$ for face, car and church images, respectively. 
% During training, all the reconstructed images and input images 
% To invert images into StyleGAN2’s latent codes, we use a pretrained
% e4e [46] as the image encoder. 
% When training our model, we leverage the text encoder
% from CLIP, and use the pre-trained StyleGAN2 to generate edited images. 
% We use 44 text prompts for face images
% containing emotion, hair color, hairstyle, age, gender, make-up, etc. Meanwhile, we follow the text
% prompts from StyleCLIP [33] when editing LSUN cars and church datasets. We randomly choose a
% text prompt for an input image for model training. 
In practice, we use a multi-step learning rate with an initial learning rate of $0.0005$. The Adam \cite{kingma2014adam} optimizer is utilized with $\beta_1$ and $\beta_2$ set to $0.9$ and $0.999$, respectively. We train our model on a single Nvidia Telsa V100 GPU. For FFHQ dataset, the total training iterations are 400,000 and the batch size is 8. For LSUN church dataset, the total training iterations are 600,000 and the batch size is 16. For LSUN cars dataset, the total training iterations are 300,000 and the batch size is 16.

The mapping network consists of $L$ fully-connected networks converts the $512$-dimensional CLIP embedding input into $L\times C$-dimensional latent codes corresponding to the $L$ layers of StyleGAN. These networks correspond to different semantic levels of information (coarse, medium and fine). For CLIP, we follow StyleCLIP to use the ViT-B/32 model to extract embeddings.

\begin{figure*}[t]
    \centering
    \includegraphics[width=\linewidth]{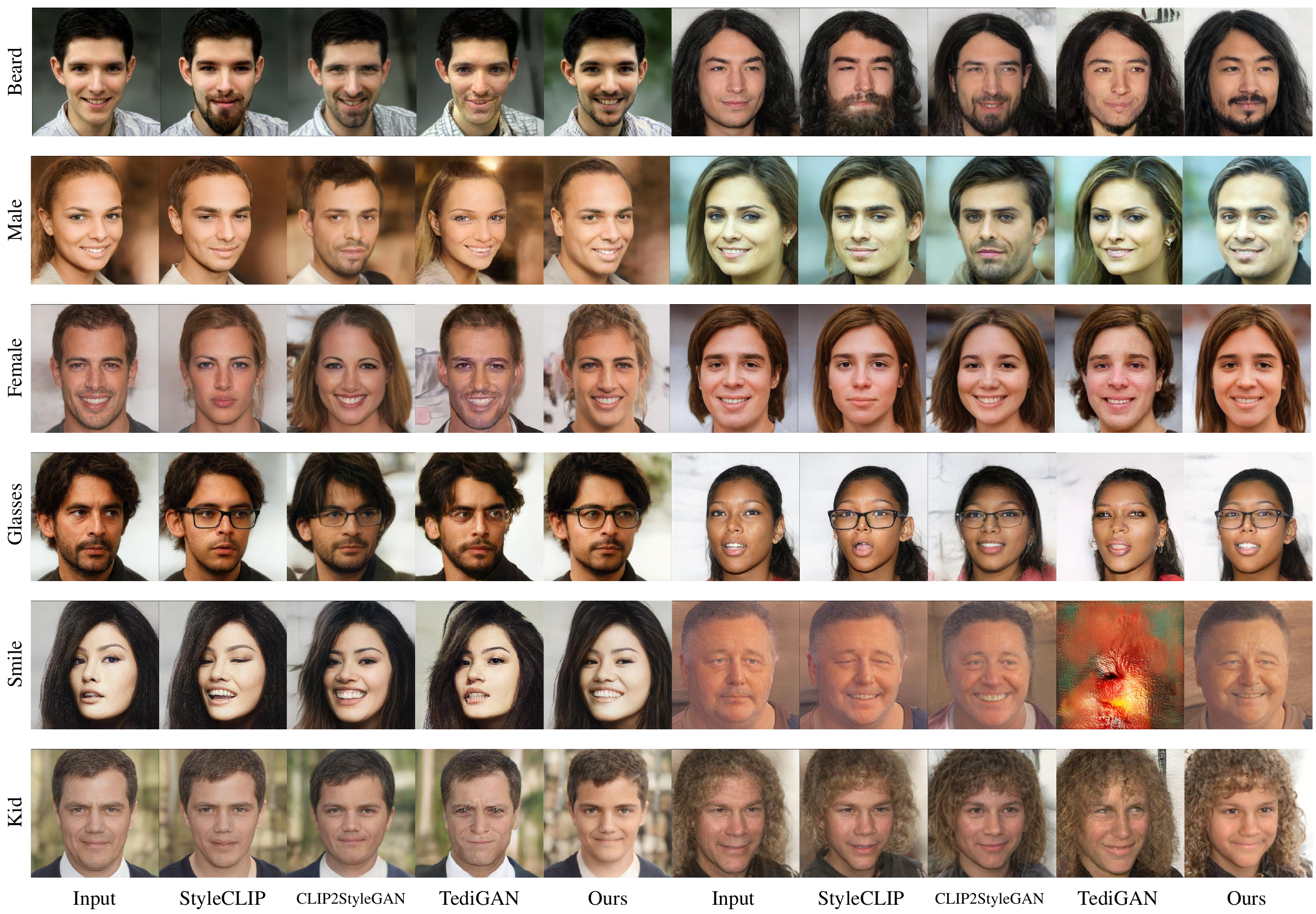}
    \caption{ Comparison results with previous text-based image manipulation methods. The previous method can lead to changes in other attributes or character identities when editing images. TediGAN even cannot edit the relevant attributes. In contrast, our method performs excellent editing effects that satisfy the text description without affecting other attributes or the identity, providing a more effective image editing tool.}
    \label{fig:com}   
    % \vspace{-0.1in}
\end{figure*}

% 之前的方法在编辑图像时会导致其他属性的变化或者任务身份的改变，甚至有一些无法做出对应属性的编辑。
% 相比之下我们的方法在很好地做满足文本描述的属性编辑的同时不会影响其他的属性或者人地身份，提供了一种更加有效地编辑方式

\subsection{Qualitative Evaluation}
We first qualitatively compare our method with several previous text-driven image manipulation methods. These methods include making attribute edits in the $W^+$ or $S$ space of StyleGAN.

%先写一下其他方法的缺点，StyleCLIP会带来其他属性的变化，消融消失，眼睛闭合
% CLIP2StyleGAN对原图的id影响很大
%TediGAN则基本对原图的属性改变很小
%都指出来问题出现在图5中的哪个位置
%再写我们的优点 ：我们的方法能很好地根据文本编辑对应的属性，而且不会影响原图的id 和其他的属性，体现了我们的方法的优越性

\textbf{Comparison with $W^+$ space methods.} These methods include TediGAN, CLIP2StyleGAN and StyleCLIP's mapper method. We compare these methods with the 6 properties in CLIP2StyleGAN. For StyleCLIP, we retrain 6 corresponding mapper networks. The comparison results are shown in Figure \ref{fig:com}. StyleCLIP fails in disentangling the attributes, i.e. when changing one attribute of an image, other attributes are also affected. For example, in the left part of Figure \ref{fig:com}, when StyleCLIP edits the ``Female'' attribute, the person's smile disappears. The person's eyes are closed when the ``Smile'' attribute is edited. CLIP2StyleGAN changes the person identity heavily when manipulating images. When CLIP2StyleGAN edits the ``Glasses'' attribute, the appearance of the person is changed, resulting in a completely different person. TediGAN hardly changes the attributes of images. When editing the ``Female'' attribute, TediGAN just adds a little shadow behind the head of the person, which is not an obvious edition. When editing the ``Glasses'' attribute, TediGAN doesn't generate the outlines of the glasses. In contrast, our method enables more accurate attributes editing while preserving other attributes and identities of images. These results demonstrate the accuracy and effectiveness of our method in using text guidance to edit images.
% In the left part of Figure \ref{fig:com}, when editing the “Smile” attribute, either other attributes or identity of the original image is changed by other methods, but our method could still produce strong results.

\begin{figure}[t]
    \centering
    \includegraphics[width=\linewidth]{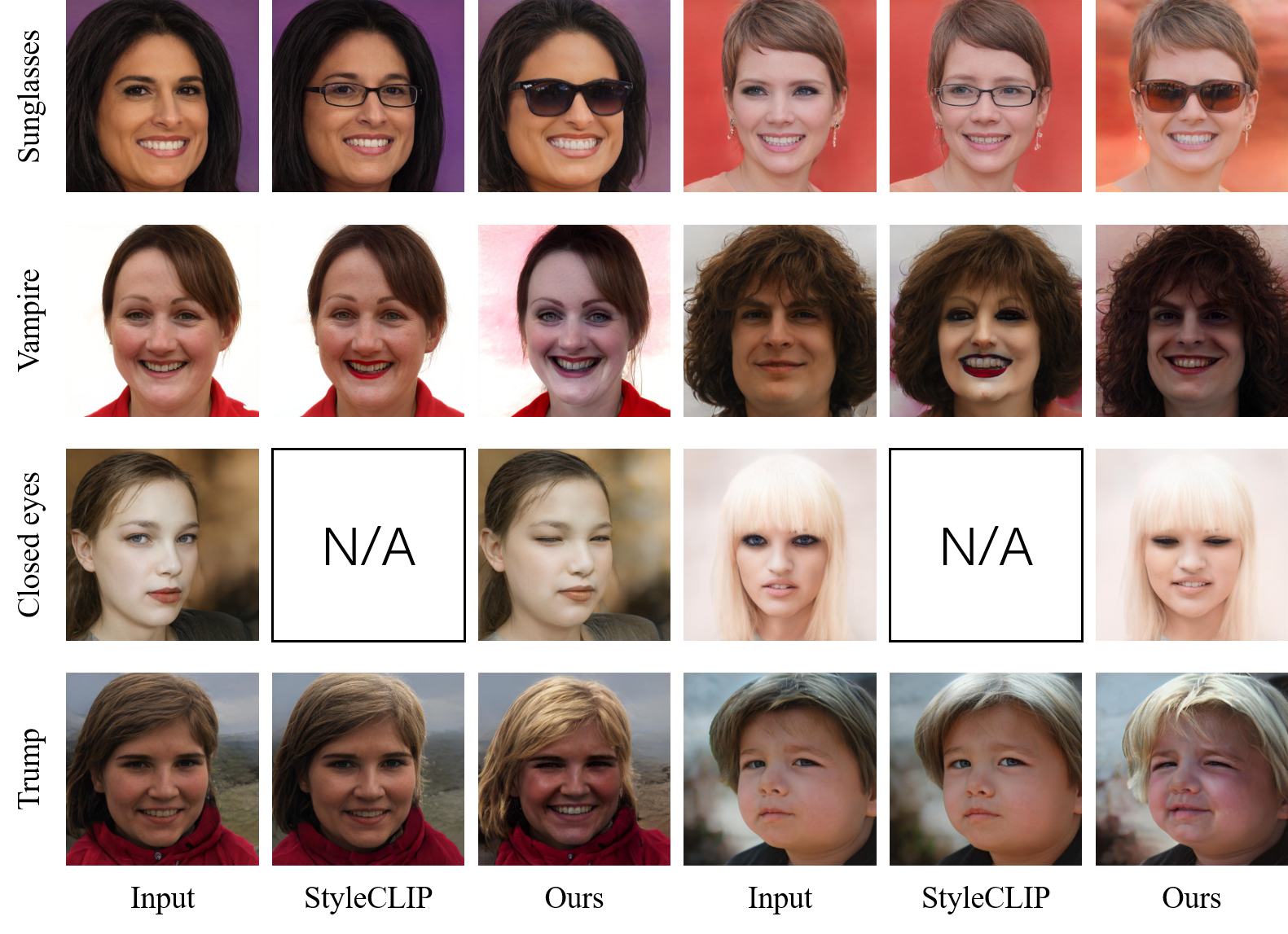}
    \caption{ Comparison with the global direction method. }
    \label{fig:global}   
    % \vspace{-0.1in}
\end{figure}

\textbf{Comparison with $S$ space method.}
We further compare our method with StyleCLIP's global direction method. Figure \ref{fig:global} shows a comparison of some abstract or detailed attributes editing effects.
%styleclip无法根据一些抽象的词汇（Vampire 和 Trump）做对应的，改变的属性很少
%对于一些细节的（sunglasses 和 closed eyes）styleclip编辑错误或者无法获得结果
StyleCLIP cannot get abstract attribute editing directions corresponding accurately. For example, in the left part of Figure \ref{fig:global}, when StyleCLIP edits the ``Vampire'' attribute, the result is almost the same as the original image. The similar situation happens when the ``Trump'' attribute is edited. For some detailed attributes, StyleCLIP generates wrong or even meaningless results. When StyleCLIP edits the ``Sunglasses'' attribute, the wrong type of glasses are generated. And StyleCLIP does not produce meaningful results when editing the ``Closed eyes'' attribute. In contrast, our method can be applied to a wider range of text conditions with higher application values.

\subsection{Quantitative Evaluation}

It is difficult to find a straight quantitative metric to evaluate the image editing results. However, we believe that the edited results should match the text prompt semantics on the one hand, and the remaining attributes (identity) do not change on the other hand. 
Therefore, we evaluate the results of these different methods from 3 perspectives: user evaluation, identity preservation, and classification accuracy.

% We conduct human subjective evaluations on the manipulated results from compared methods.
\textbf{User evaluation.} 
We randomly collect 108 images which are edited by the above 6 text prompts to perform the user evaluation. 35 participants from different backgrounds participate in this evaluation. They are asked to observe the editing results of the different methods and select the image that better matches the text description with better identity preservation. Each participant is asked to complete all choices within 20 minutes. The evaluation scores are shown in Table \ref{tab:user}. The results of our method are mostly preferred.

%35名不同背景的参与者被要求根据三个同样重要的原则投票选出最佳结果。首先，他们应该选择语义最符合文本提示的结果。其次，他们应该选择最能保留人类身份的结果。
% 35名不同背景的参与者参与了该评估。他们被要求观察不同方法的编辑结果，并选择更符合文本描述、与原图像身份更一致的结果的图像。
% 被要求在所有方法中选择其中符合文本描述并且原图像身份更一致的结果

% \begin{tabular}{c|ccc|ccc}
% \hline \multirow{2}{*}{ Text Prompt } & \multicolumn{3}{|c|}{ Editing Performance } & \multicolumn{2}{c}{ Human } & Subject Evaluation \\
% \cline { 2 - 7 } & TediGAN & StyleCLIP & Ours & TediGAN & StyleCLIP & Ours \\
% \hline Bald & $0.2507$ & $0.1015$ & $\mathbf{0 . 0 2 7 9}$ & $5.7 \%$ & $8.6 \%$ & $\mathbf{8 5 . 7} \%$ \\
% Angry & $0.4520$ & $0.4860$ & $\mathbf{0 . 5 8 1 0}$ & $0.0 \%$ & $17.1 \%$ & $\mathbf{8 2 . 9} \%$ \\
% Red hair & $1.0971$ & $1.0416$ & $\mathbf{0 . 7 1 7 1}$ & $2.9 \%$ & $0.0 \%$ & $\mathbf{9 7 . 1} \%$ \\
% Aged 10 & $21.1448$ & $17.0053$ & $\mathbf{9 . 8 8 7 4}$ & $2.9 \%$ & $0.0 \%$ & $\mathbf{9 7 . 1} \%$ \\
% Bowl cut hairstyle & $-$ & $-$ & $-$ & $2.9 \%$ & $5.7 \%$ & $\mathbf{9 1 . 4} \%$ \\
% Beard blond mohawk hair & $-$ & $-$ & $-$ & $8.6 \%$ & $0.0 \%$ & $\mathbf{9 1 . 4} \%$ \\
% \hline
% \end{tabular}

\begin{table}
    \small
    \centering 
    \caption{The user evaluation scores for different methods. The higher the better. Among these methods, our results have the highest preference.}
\begin{tabular}{c|c|c|c|c} 
\hline
\multirow{2}{*}{ Text } & \multicolumn{4}{c}{ User Evaluation } \\
\cline { 2 - 5 } & TediGAN & StyleCLIP& CLIP2StyleGAN & Ours \\
\hline \hline Beard & \makecell[c]{2.8\%} & \makecell[c]{24.9\%}& \makecell[c]{16.6\%}& \makecell[c]{\textbf{55.7}\%} \\
Male & \makecell[c]{0\%}  & \makecell[c]{19.6\%}& \makecell[c]{13.9\%}& \makecell[c]{\textbf{66.5}\%} \\
Female & \makecell[c]{0\%}& \makecell[c]{18.1\%}& \makecell[c]{16.6\%}& \makecell[c]{\textbf{65.3}\%} \\
Glasses & \makecell[c]{0\%}& \makecell[c]{26.6\%}& \makecell[c]{20.7\%}& \makecell[c]{\textbf{52.7}\%} \\
Smile & \makecell[c]{0\%}& \makecell[c]{28.9\%}& \makecell[c]{12.1\%}& \makecell[c]{\textbf{59}\%} \\
kid & \makecell[c]{4.2\%}& \makecell[c]{20.7\%}& \makecell[c]{10.7\%}& \makecell[c]{\textbf{64.4}\%} \\
\hline
\end{tabular}
    \label{tab:user}
\end{table}

% \textbf{Attribute analysis.}
% We utilize several text prompts and randomly select 1000
% 234 testing images from the CelebA-HQ dataset. 
% Then, we predict the attribute
\textbf{ID preservation analysis.} We use ArcFace \cite{deng2019arcface} to extract embedding vectors of generated images to compare identity preservation to other methods.
We use a total of 1000 images edited with six attributes to perform this evaluation.  The average scores of face identity for different edited results are shown in Table \ref{tab:id}. Among all the methods, our method can maintain the face identity best.

\textbf{CLIP classification accuracy.} We use CLIP as a classifier to evaluate whether the edited results have the target attributes. Texts with and without corresponding attribute descriptions are used as labels for binary classification. The average accuracy of classification is shown in Table \ref{tab:clip}.
The highest accuracy rate proves that our method works better for editing target attributes.

\begin{table}
    \small
    \centering 
    \caption{ID preservation scores of different methods.}
\begin{tabular}{c|c|c|c|c} 
\hline
\multirow{2}{*}{ Text } & \multicolumn{4}{c}{ ID Score } \\
\cline { 2 - 5 } & TediGAN & StyleCLIP& CLIP2StyleGAN & Ours \\
\hline \hline Beard & \makecell[c]{0.79} & \makecell[c]{0.72}& \makecell[c]{0.58}& \makecell[c]{\textbf{0.89}} \\
Male & \makecell[c]{0.81} & \makecell[c]{0.69}& \makecell[c]{0.49}& \makecell[c]{\textbf{0.86}} \\
Female & \makecell[c]{0.78} & \makecell[c]{0.68}& \makecell[c]{0.51}& \makecell[c]{\textbf{0.84}} \\
Glasses & \makecell[c]{0.76} & \makecell[c]{0.71}& \makecell[c]{0.54}& \makecell[c]{\textbf{0.88}} \\
Smile & \makecell[c]{0.81} & \makecell[c]{0.78}& \makecell[c]{0.61}& \makecell[c]{\textbf{0.92}} \\
Kid & \makecell[c]{0.72} & \makecell[c]{0.73}& \makecell[c]{0.60}& \makecell[c]{\textbf{0.76}} \\

\hline
\end{tabular}
    \label{tab:id}
\end{table}

\begin{table}
    \small
    \centering 
    \caption{The CLIP classification accuracy results.}
\begin{tabular}{c|c|c|c|c} 
\hline
\multirow{2}{*}{ Text } & \multicolumn{4}{c}{ Classification accuracy } \\
\cline { 2 - 5 } & TediGAN & StyleCLIP& CLIP2StyleGAN & Ours \\
\hline \hline Beard & \makecell[c]{0.31} & \makecell[c]{0.88}& \makecell[c]{0.86}& \makecell[c]{\textbf{0.94}} \\
Male & \makecell[c]{0.42} & \makecell[c]{0.90}& \makecell[c]{0.88}& \makecell[c]{\textbf{0.96}} \\
Female & \makecell[c]{0.55} & \makecell[c]{0.89}& \makecell[c]{0.91}& \makecell[c]{\textbf{0.97}} \\
Glasses & \makecell[c]{0.22} & \makecell[c]{\textbf{0.99}}& \makecell[c]{0.92}& \makecell[c]{\textbf{0.99}} \\
Smile & \makecell[c]{0.43} & \makecell[c]{0.87}& \makecell[c]{0.88}& \makecell[c]{\textbf{0.94}} \\
Kid & \makecell[c]{0.26} & \makecell[c]{0.82}& \makecell[c]{0.82}& \makecell[c]{\textbf{0.96}} \\

\hline
\end{tabular}
    \label{tab:clip}
\end{table}

% This is done by generating 10K image pairs that share the ID attribute and
% have different pose, illumination and expression attributes.
%一方面就是属性编辑的效果要有，另一方面就是对于没有改变id的编辑要计算id之间的差异

\subsection{Ablation Studies}
% Then, we follow [36] to use the multiple semantic classification models to
% 236 measure the text-relevance of these results.

% \begin{table}
%     \small
%     \centering 
%     \caption{The user study scores for different methods. The higher the better. Among the three methods, our results have the highest preference.}
% \begin{tabular}{l|l} 
% \hline
% Methods & Preference Score  \\
% \hline \hline CLIPstyler & \makecell[c]{12.1\%}  \\
% LDAST & \makecell[c]{4.6\%}   \\
% Ours & \makecell[c]{82.7\%} \\
% \hline
% \end{tabular}
%     \label{tab:user}
% \end{table}

% \textbf{global vs. personal}

\begin{figure}[t]
    \centering
    \includegraphics[width=\linewidth]{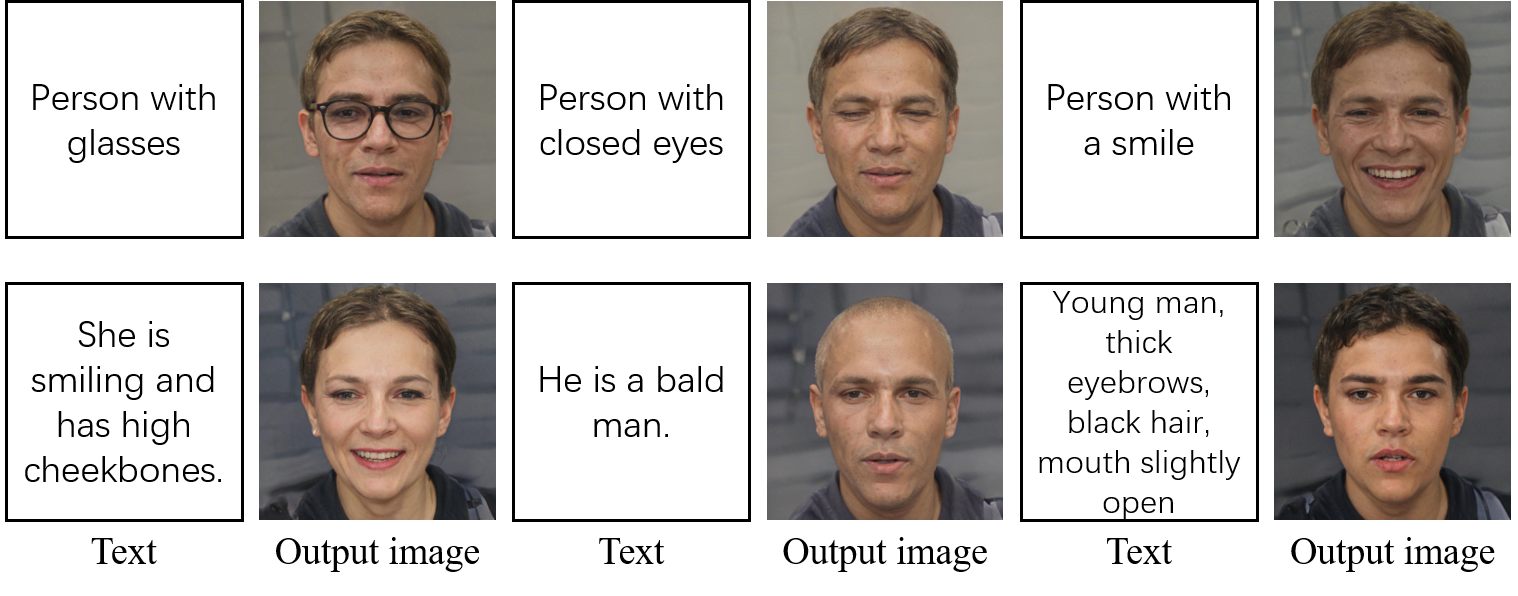}
    \caption{ Text to image result. }
    \label{fig:text2image}   
    % \vspace{-0.1in}
\end{figure}

% (figure)
\textbf{Text-to-image results.}
As mentioned above, we use some general text descriptions to find arbitrary semantic shifts. These descriptions are mapped to an average face with certain attributes. In Figure \ref{fig:text2image}, we show some faces generated with text descriptions. We can see that these faces have the attributes described by the text, and the identity is consistent under different attributes.
%如上面所说的，我们使用一些general的文本描述来找到任意的语义的转换。这些描述会被映射为一个具有某种属性的平均人脸。在图7中我们展示了一些用文本描述所生成的人脸，可以看到这些人脸很好地对应了文本所描述的属性，并且在不同的属性下的身份一致

\begin{figure}[t]
    \centering
    \includegraphics[width=\linewidth]{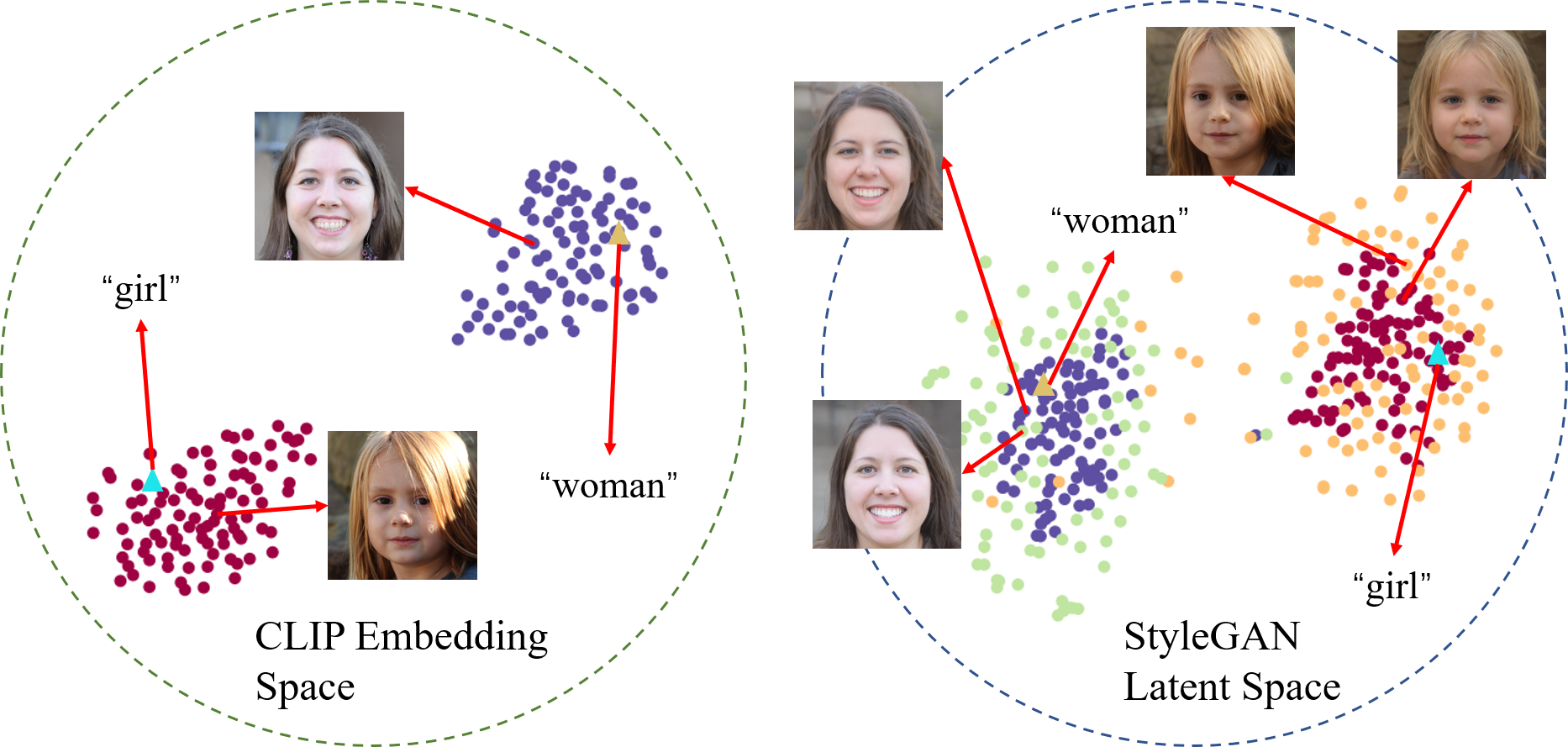}
    \caption{ Space visualization. }
    \label{fig:tsne}   
    % \vspace{-0.1in}
\end{figure}

\textbf{Space visualization.}
To better display the accuracy of our Space Alignment module, we make a visualization for the two spaces. We use 100 images of women and 100 images of girls for the analysis. These 200 images are encoded using CLIP's encoder, e4e's encoder, and our network to obtain the corresponding latent representations. We then visualize these codes using t-SNE, and the result is shown in Figure \ref{fig:tsne}. The blue and red dots in the StyleGAN space represent the corresponding codes mapped from the CLIP space. The codes of the text are represented using triangular symbols that fall between the image codes of the corresponding attributes. It can be seen that our method is able to map the text embeddings to the exact semantic positions in StyleGAN space.
%为了更好的展示我们方法的空间对齐的效果，我们对于两个空间做了一个可视化展示.我们使用100张女人的图像和100张女孩的图像进行分析，这200张图像分别使用clip的encoder编码，使用e4e的编码器编码，使用我们的网络映射到stylegan空间，得到对应的编码表示。然后我们对于这些编码进行使用tsne进行可视化，结果如图8所示，stylegan空间中的蓝点和红点表示对应的从clip空间映射过来的code其余的圆点则是通过e4e获得的code。文本所对应的code使用了三角符号表示，落在相应属性的图像编码之间。可以看到我们的方法能够很准确地将将文本的编码映射到stylegan空间中的对应位置上去

\begin{figure}[t]
    \centering
    \includegraphics[width=\linewidth]{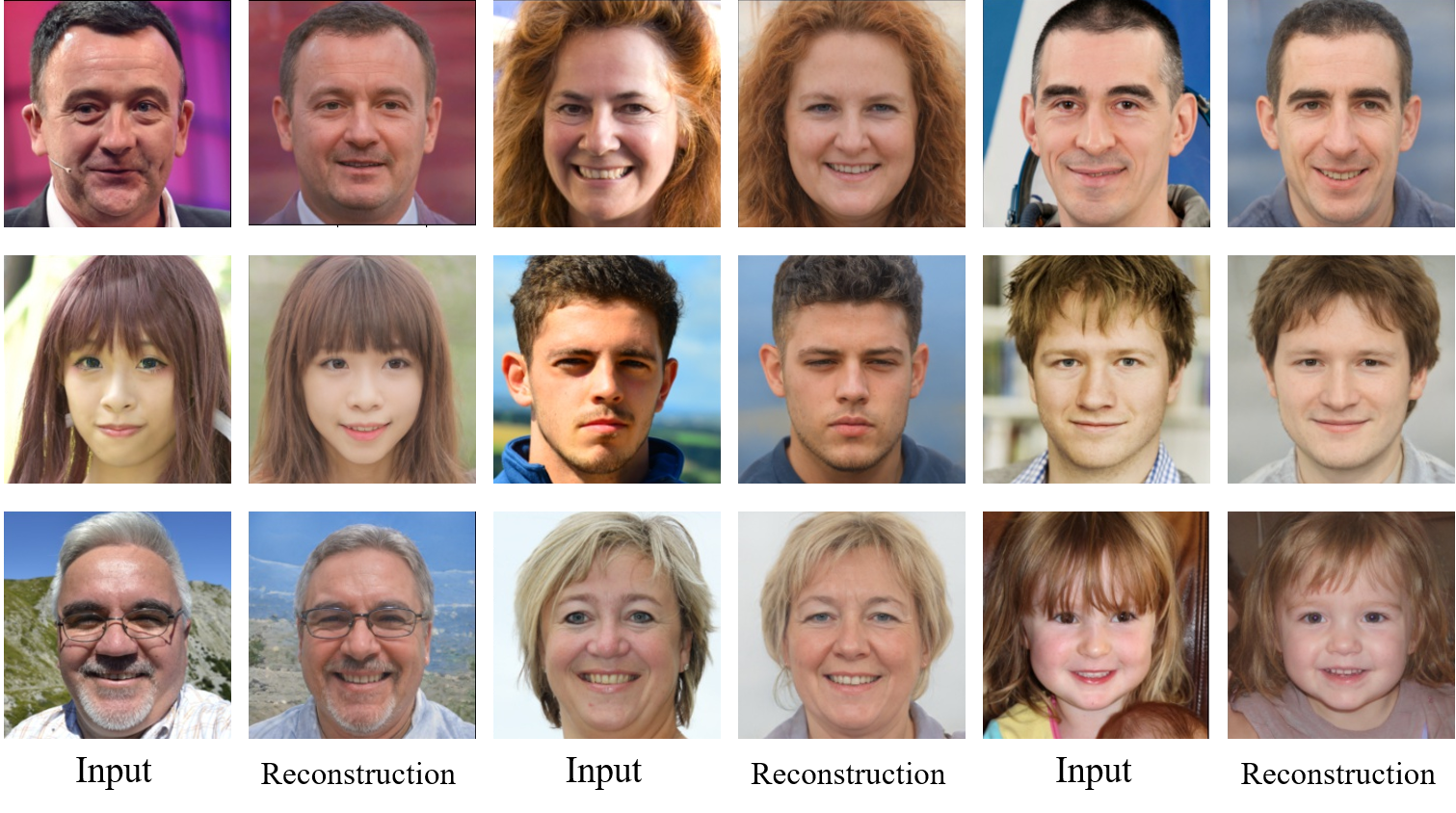}
    \caption{ Reconstruction results. }
    \label{fig:reconstruction}   
    % \vspace{-0.1in}
\end{figure}

\textbf{Reconstruction results.}
The training objective of our method is to use the latent codes mapped from CLIP's image embeddings to generate results with the same attributes as the original image. 
In Figure \ref{fig:reconstruction}, we show some image reconstruction results obtained in this training method. It can be seen that these results are consistent with the original image in terms of gender, age, expression and hairstyle, etc.
%如上面所说的，我们的方法的目标是得到有相同属性的图像而不是做完全一致的图像
%可以看到我们的方法保证了性别，年龄表情，发型等属性一致

\begin{figure}[t]
    \centering
    \includegraphics[width=\linewidth]{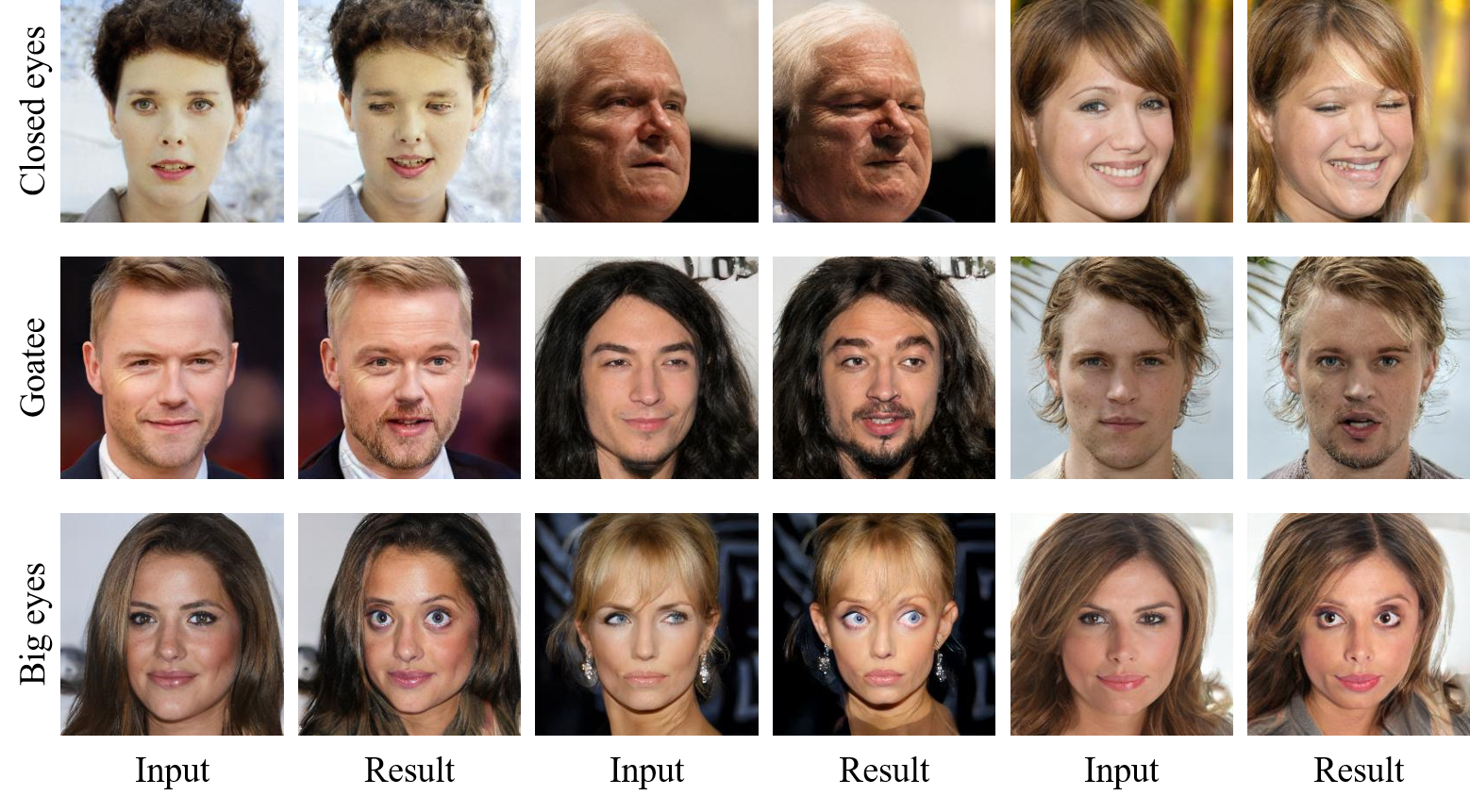}
    \caption{ Combination with the HFGI. }
    \label{fig:HFGI}   
    % \vspace{-0.1in}
\end{figure}

\begin{figure}[t]
    \centering
    \includegraphics[width=\linewidth]{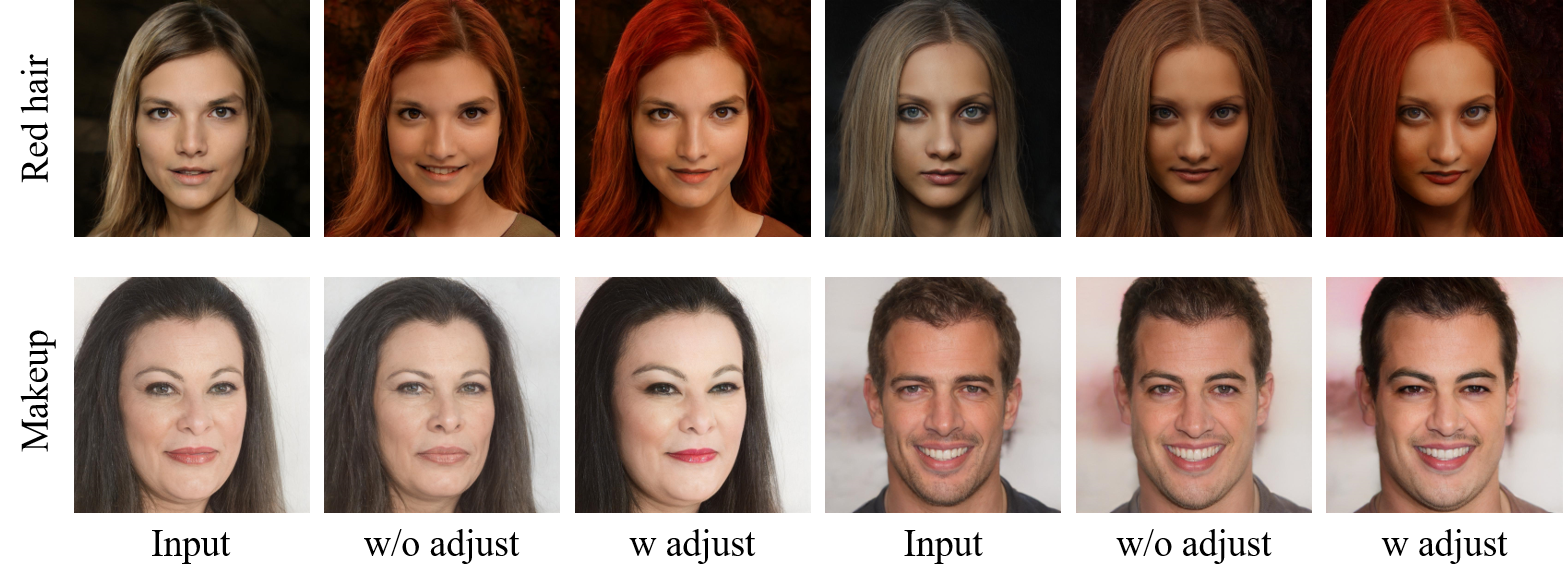}
    \caption{ Ablation study of In-domain Adjustment. }
    \label{fig:adjust}   
    % \vspace{-0.1in}
\end{figure}

\textbf{Combination with other inversion method.}
Our semantic editing method can be combined with other inversion methods without adaptation. To illustrate this, we combine our method with a recent high-fidelity inversion method HFGI\cite{wang2022high} without adaption. The relevant editing results are shown in Figure \ref{fig:HFGI}.
These edited results also match the text descriptions, demonstrating the robustness of our method.
% Meanwhile, as shown in Fig. 21, although the FFCLIP is trained with e4e
% 502 encoder, it can edit image correct with High-fidelity [43] and Restyle [5] encoders, respectively. These
% 503 phenomenons prove the robustness of our model and the effectiveness of the semantic alignment
% 504 module.

\begin{figure}[t]
    \centering
    \includegraphics[width=\linewidth]{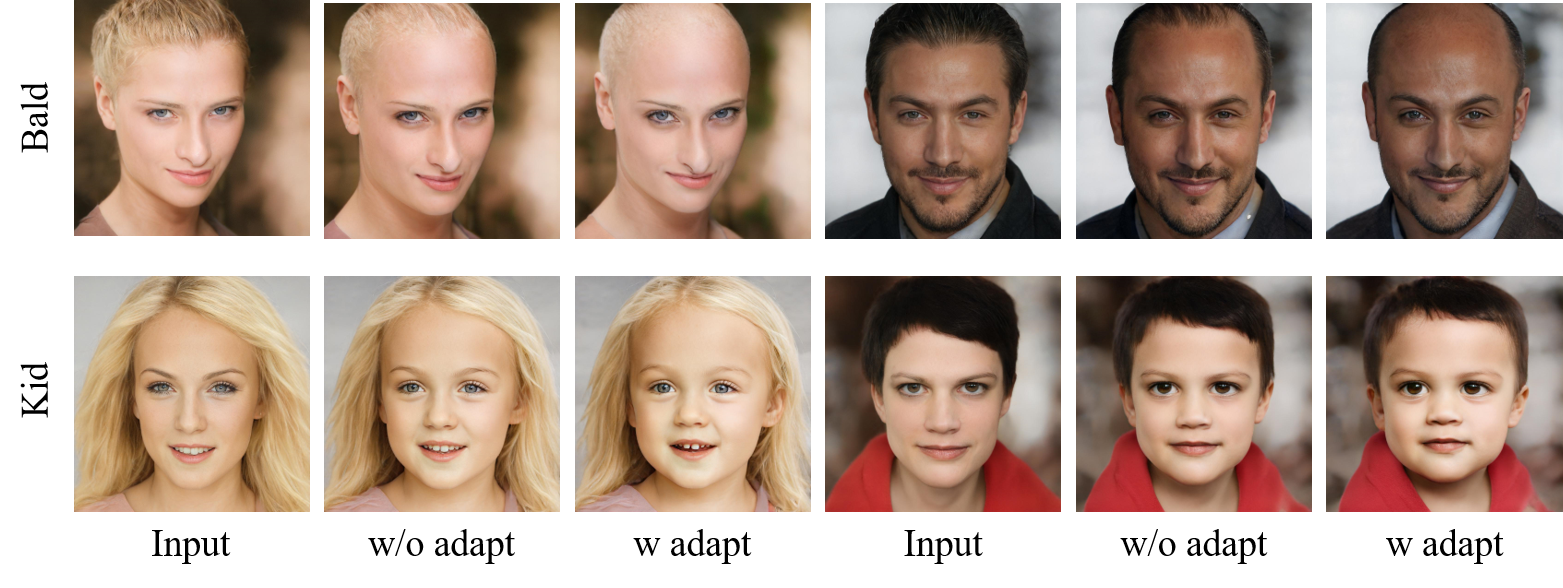}
    \caption{Ablation study of Adaption. }
    \label{fig:adapt}   
    % \vspace{-0.1in}
\end{figure}

\textbf{Alignment process.}
We also explore the effect of each part of our method. To demonstrate the effect of our In-domain Adjustment, we eliminate this process and retrain a mapping network using only real image data, also without adaption. See Figure \ref{fig:adjust} for a comparison of the editing results using this network and the network trained with the generated data. After using the generated image for adjustment, the edited attributes are more accurate, such as redder hair color and more obvious makeup. We further explore the effect of the specific adaption when combining our method with other inversion schemes. We use networks with and without adaption to e4e's encoder to edit the images. Figure \ref{fig:adapt} shows the comparison results. Some details are more perfect when adapting the network to e4e.
% It can be seen that the editing results of our network without specific adaption to e4e are also accurate. However, 
%我们也探究了在与其他的反演方法结合又有对其坐骑做特定的调整的影响。我们分别用有和没有与e4e的编码器相适应的网络对图像做了对图像进行编辑，相关的对比结果展示在图10中。可以看到在没有与e4e做特定的调整的时候，我们方法编辑的效果也是准确的，不过在于e4e相适应后一些细节的效果会更加准确

%在这里我们进一步探究了我们方法的效果，为了证明我们indomain调整的效果，我们消除了这一过程重新只用真实图像数据重新训练一个映射网络，也没有针对于e4e进行调整。使用这个网络以及使用生成训练的数据获得网络的效果的对比见图11

% 有没有加生成图像的对比 有没有针对于e4e的特征空间overfit的一个对比   

%最后加一个两个消融的结果
%text2image的效果
%还有一个表

% 以及和restyle的效果
%global和personal的看一下吧
%reconstruction看一下吧

\section{Conclusion}

%在这个工作里我们提出里一个新的技术叫做空间对齐，目标是将预训练好的语言图像嵌入空间和生成模型的隐式空间相同语义的区域对齐.在完成对齐后，我们可以使用文本描述在stylegan的空间中找到具有对应属性的latent。这些latent被用来抽取任意语义的语义编辑用于编辑stylegan的图像。相比之前的编辑的方法，我们的方法不需要手动标注以及大量的计算资源 我们的方法为用户提供了一个灵活的接口来编辑各种不同的属性，为生成模型的隐式空间的语义探索提供了一个新的思路

In this work we propose a new image manipulation framework TMSA based on Space Alignment. Space Alignment aims to align the same sematic regions in pre-trained language-image embedding space and the latent space of StyleGAN. After the alignment is done, we can use text descriptions to find latent codes with corresponding attributes in the StyleGAN space. Then, these latent codes can be used to extract semantic shift of arbitrary semantics for editing the imagery of StyleGAN. Compared to previous editing methods, our method does not require manual efforts for extensive annotation or time-consuming training. Our method provides a flexible interface for user to edit different attributes of input image and is a new way to explore the semantics in the latent space of generative model.

% Update the cvpr.cls to do the following automatically.
% For this citation style, keep multiple citations in numerical (not
% chronological) order, so prefer \cite{Alpher03,Alpher02,Authors14} to
% \cite{Alpher02,Alpher03,Authors14}.

% \begin{figure*}
%   \centering
%   \begin{subfigure}{0.68\linewidth}
%     \fbox{\rule{0pt}{2in} \rule{.9\linewidth}{0pt}}
%     \caption{An example of a subfigure.}
%     \label{fig:short-a}
%   \end{subfigure}
%   \hfill
%   \begin{subfigure}{0.28\linewidth}
%     \fbox{\rule{0pt}{2in} \rule{.9\linewidth}{0pt}}
%     \caption{Another example of a subfigure.}
%     \label{fig:short-b}
%   \end{subfigure}
%   \caption{Example of a short caption, which should be centered.}
%   \label{fig:short}
% \end{figure*}

%------------------------------------------------------------------------

%-------------------------------------------------------------------------

%-------------------------------------------------------------------------

%-------------------------------------------------------------------------

%-------------------------------------------------------------------------

%-------------------------------------------------------------------------

%-------------------------------------------------------------------------

%------------------------------------------------------------------------

%%%%%%%%% REFERENCES
{\small
\bibliographystyle{ieee_fullname}
\bibliography{egbib}
}

\end{document}